\documentclass{article}

\usepackage{PRIMEarxiv}

\usepackage[utf8]{inputenc} 
\usepackage[T1]{fontenc}    
\usepackage{hyperref}       
\usepackage{url}            
\usepackage{booktabs}       
\usepackage{amsfonts}       
\usepackage{nicefrac}       
\usepackage{microtype}      
\usepackage{lipsum}
\usepackage{fancyhdr}       
\usepackage{graphicx}   
\usepackage{float}
\graphicspath{{media/}}     

\title{NordFKB: a fine-grained benchmark dataset for geospatial AI in Norway}

\author{
  Sander Riisøen Jyhne \\
  Kartverket \\
  Kristiansand\\
  \texttt{sander.jyhne@kartverket.no} \\
  \And
  Aditya Gupta \\
  University of Agder \\
  Grimstad\\
  \texttt{aditya.gupta@uia.no} \\
  \And
  Ben Worsley \\
  Kartverket \\
  Kristiansand\\
  \texttt{ben.worsley@kartverket.no} \\
  \And
  Marianne Andersen \\
  Kartverket \\
  Kristiansand\\
  \texttt{marianne.andersen@kartverket.no} \\
  \And
  Ivar Oveland \\
  Kartverket \\
  Kristiansand \\
  \textit{ivar.oveland@kartverket.no}
  \And
  Alexander Salveson Nossum \\
  Norkart \\
  Kristiansand \\
  \textit{alexander.nossum@norkart.no}
}

\begin{document}
\maketitle

\begin{abstract}
We present NordFKB, a fine-grained benchmark dataset for geospatial AI in Norway, derived from the authoritative, highly accurate, national Felles KartdataBase (FKB). The dataset contains high-resolution orthophotos paired with detailed annotations for 36 semantic classes, including both per-class binary segmentation masks in GeoTIFF format and COCO-style bounding box annotations. Data is collected from seven geographically diverse areas, ensuring variation in climate, topography, and urbanization. Only tiles containing at least one annotated object are included, and training/validation splits are created through random sampling across areas to ensure representative class and context distributions. Human expert review and quality control ensures high annotation accuracy. Alongside the dataset, we release a benchmarking repository with standardized evaluation protocols and tools for semantic segmentation and object detection, enabling reproducible and comparable research. NordFKB provides a robust foundation for advancing AI methods in mapping, land administration, and spatial planning, and paves the way for future expansions in coverage, temporal scope, and data modalities.
\end{abstract}

\section*{Keywords}
Geospatial AI, Aerial Imagery, Semantic Segmentation, Object Detection, Benchmark Dataset

\section{Introduction}

Benchmark datasets are critical for advancing artificial intelligence (AI), particularly in domains where general-purpose datasets fail to capture local geographic and semantic complexity. In geospatial AI, high-resolution, locally consistent data is essential for training, evaluating, and comparing models. Yet, despite Norway’s extensive national mapping infrastructure, openly available benchmark datasets based on high-resolution (<15 cm) aerial imagery remain scarce. This is largely due to the restricted availability of orthophotos and vector data, with only select regions, such as those used in NordFKB, released for open use.

The Norwegian National Mapping Database (FKB) is one of the world’s most detailed, standardized, and semantically rich geospatial resources. Maintained through coordinated national mapping efforts, it contains precise representations of buildings, terrain, infrastructure, and land cover \cite{geovekst}. Despite this richness, FKB has so far been underutilized as a foundation for AI research. Prior efforts such as the MapAI competition \cite{mapai}, part of the KartAi initiative \cite{kartai}, demonstrated the value of structured national data but were limited in scope, focusing solely on building segmentation. Across the KartAi project series, a consistent conclusion has emerged: there is a lack of generally available, high-quality benchmark datasets for AI in the Norwegian geospatial domain \cite{ab2_bygg, optimalisering_ai_analyse, pålitelighetsmål, kartai_effektivisering, automatisering, tilgjengeliggjøring, kvalitetskontroll, bygningskontroll}.

This work introduces \textbf{NordFKB}, a fine-grained benchmark dataset that transforms FKB-derived geospatial data into a standardized, machine-learning-ready resource supporting 36 semantic classes across diverse Norwegian landscapes. NordFKB is designed to support a broad range of geospatial AI tasks, such as semantic segmentation and object detection, while promoting methodological comparability through consistent annotation standards and evaluation protocols. By making high-quality, nationally sourced geospatial data openly accessible, NordFKB aims to lower the barrier to entry for geospatial AI research in Norway, enable fair model comparisons, and encourage the development of methods that generalize across heterogeneous geographic and environmental conditions.

\section{Related Work}

Several international geospatial AI benchmarks have been introduced in recent years, each addressing different tasks and geographic settings. The SpaceNet competition series \cite{spacenet} has been a major driver in the field, offering public datasets for building footprint and road extraction from high-resolution satellite imagery. Primarily focused on U.S. urban areas, it helped establish common evaluation protocols and demonstrated the value of open, reproducible benchmarks for remote sensing tasks. However, its geographic scope and limited class diversity reduce its applicability for fine-grained mapping in a Norwegian context, where national mapping standards and feature definitions differ.

The DeepGlobe challenge \cite{deepglobe} broadened the scope by targeting more varied environments, including rural and agricultural regions. The challenge included benchmarks for road extraction, building detection, and land cover classification, drawing attention to the difficulty of building models that generalize across diverse geographic domains. While offering greater geographic variety, its coarser class definitions and lack of authoritative mapping data limit its relevance for national mapping workflows.

The INRIA Aerial Image Labeling dataset \cite{inria} focused on cross-region generalization in binary building segmentation, using aerial imagery from cities in Europe and the United States. Although widely adopted, its scope is narrow and includes only a single object class, offering no support for multi-class or fine-grained feature analysis.

Beyond these early benchmarks, several recent datasets have targeted broader semantic coverage or more diverse geographies. LoveDA \cite{Wang2021_LoveDA} provides semantic segmentation labels across urban and rural environments in China, offering domain-diverse scenes but relying on non-authoritative annotations and limited class definitions. BigEarthNet \cite{Sumbul2019_BigEarthNet} introduces large-scale multi-label scene classification from Sentinel-2 imagery, valuable for remote-sensing research but too coarse in resolution and semantics for detailed mapping tasks. The IEEE GRSS Data Fusion Contest 2020 (DFC2020) \cite{Yokoya2020_DFC2020} combines airborne hyperspectral, LiDAR and RGB data for semantic segmentation, but focuses on multisensor fusion rather than authoritative cartographic features. More recently, OpenEarthMap \cite{Xia2023_OpenEarthMap} offers globally distributed, manually annotated building footprints and land-cover classes, providing geographic diversity but without alignment to national mapping specifications.

In a Nordic context, the MapAI competition \cite{mapai} introduced a dataset based on authoritative mapping data from Norway and Denmark. This work demonstrated the value of government-produced geospatial data for building segmentation using orthophotos and rasterized LiDAR. However, it was framed as a competition with a single task and limited semantic scope.

NordFKB builds on the foundation of earlier work and extends it into a fine-grained benchmark with 36 semantic classes, covering a broad range of geographic features beyond buildings. Combined with a standardized evaluation protocol and an accompanying benchmarking repository, it is designed to support long-term geospatial AI development and enable more challenging tasks such as multi-class object detection and segmentation.

\section{Dataset Sources and Collection}

FKB is a collection of primary geospatial datasets jointly collected and maintained by municipalities and partners in the Geovekst collaboration \cite{geovekst}. Geovekst is a national collaboration and standardization body for the establishment, quality standardization, management, and use of geographic information in Norway. The partners include the Norwegian Mapping Authority (Kartverket) \cite{kartverket}, the Norwegian Public Roads Administration (Statens Vegvesen) \cite{svv}, the Norwegian Water Resources and Energy Directorate (NVE) \cite{nve}, Bane NOR \cite{banenor}, Energi Norge, Telenor \cite{telenor}, the municipalities, the county municipalities, and the Ministry of Agriculture and Food along with its subordinate agencies.

This governance model ensures that FKB data is both authoritative, a known quality and up to date, as it is produced and validated through coordinated national mapping efforts and contributions from multiple sectoral stakeholders. As a result, FKB is considered one of the most reliable and consistent sources of detailed geographic information in Norway, making it well suited as the foundation for fine-grained AI-based mapping and analysis.

\begin{figure}[htbp]
    \centering
    \includegraphics[width=0.6\textwidth]{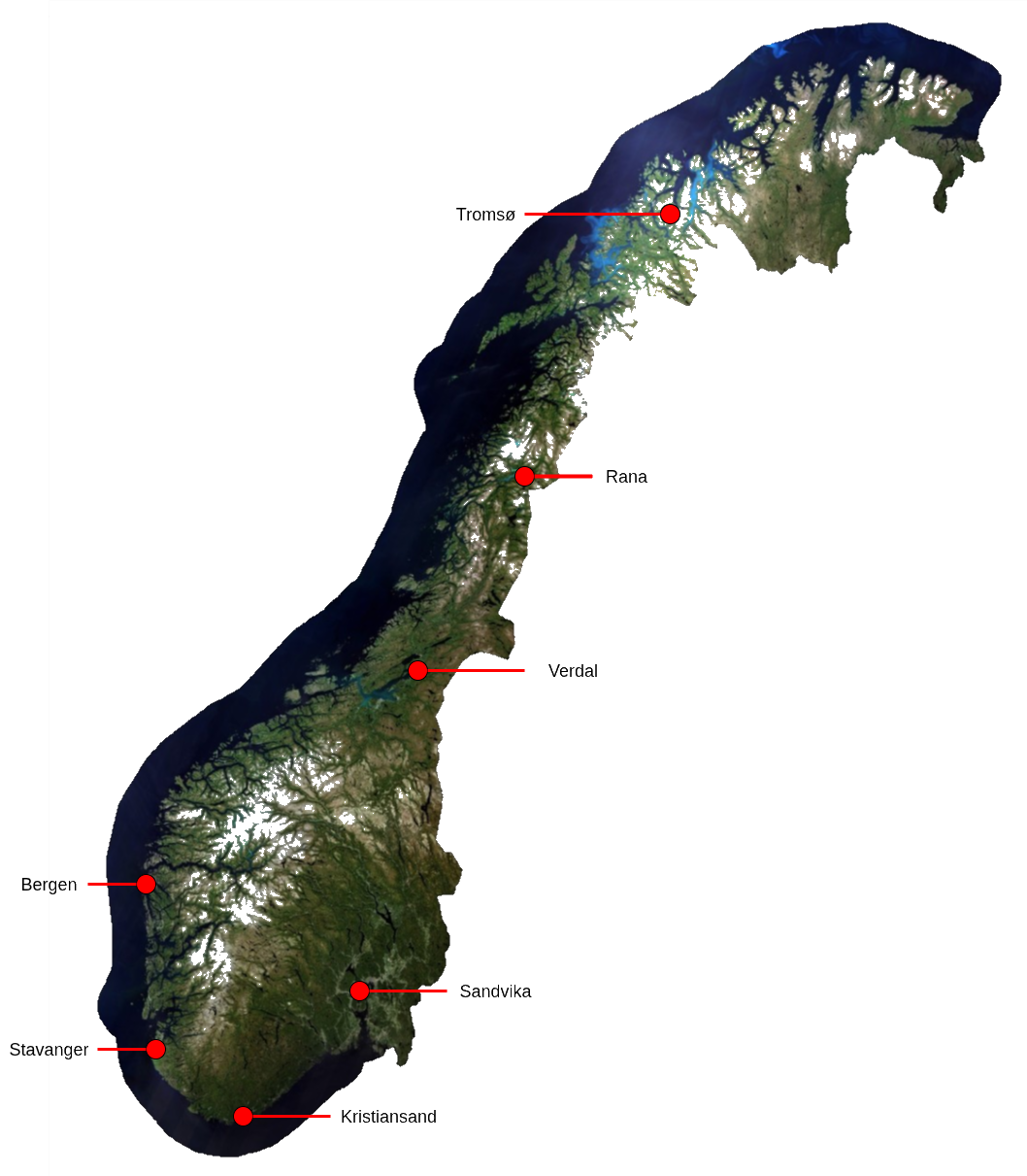}
    \caption{Overview map showing the seven geographic areas included in the NordFKB dataset: Bergen, Kristiansand, Rana, Sandvika, Stavanger, Tromsø, and Verdal. The areas are distributed across Norway, covering a range of climatic, topographic, and urbanization conditions.}

    \label{fig:norway_map}
\end{figure}

For NordFKB, we selected seven geographically diverse areas to maximize representativeness and ensure broad applicability as a benchmark for the entirety of Norway. The selected areas are Bergen, Kristiansand, Rana, Sandvika, Stavanger, Tromsø, and Verdal (see Figure~\ref{fig:norway_map}). Together they span the length of the country and capture variations in climate, topography, and urbanization.

Table~\ref{tab:area_analysis} summarizes dataset statistics for each area, including the number of images, total annotations, number of classes present, and average annotations and classes per image. These figures illustrate the variation in geographic content and annotation density across different regions, which supports evaluating model robustness under varying conditions.

The benchmark includes orthophotos in combination with bounding boxes and segmentation masks for 36 distinct geographic feature classes derived from the FKB database. All dataset components have undergone manual visual quality control to ensure accuracy and consistency. Any detected errors or inconsistencies during this process have been corrected, resulting in a benchmark dataset with a high level of reliability for both research and practical applications.

\begin{figure}[H]
    \centering
    \begin{minipage}[t]{0.45\textwidth}
        \centering
        \includegraphics[width=\textwidth]{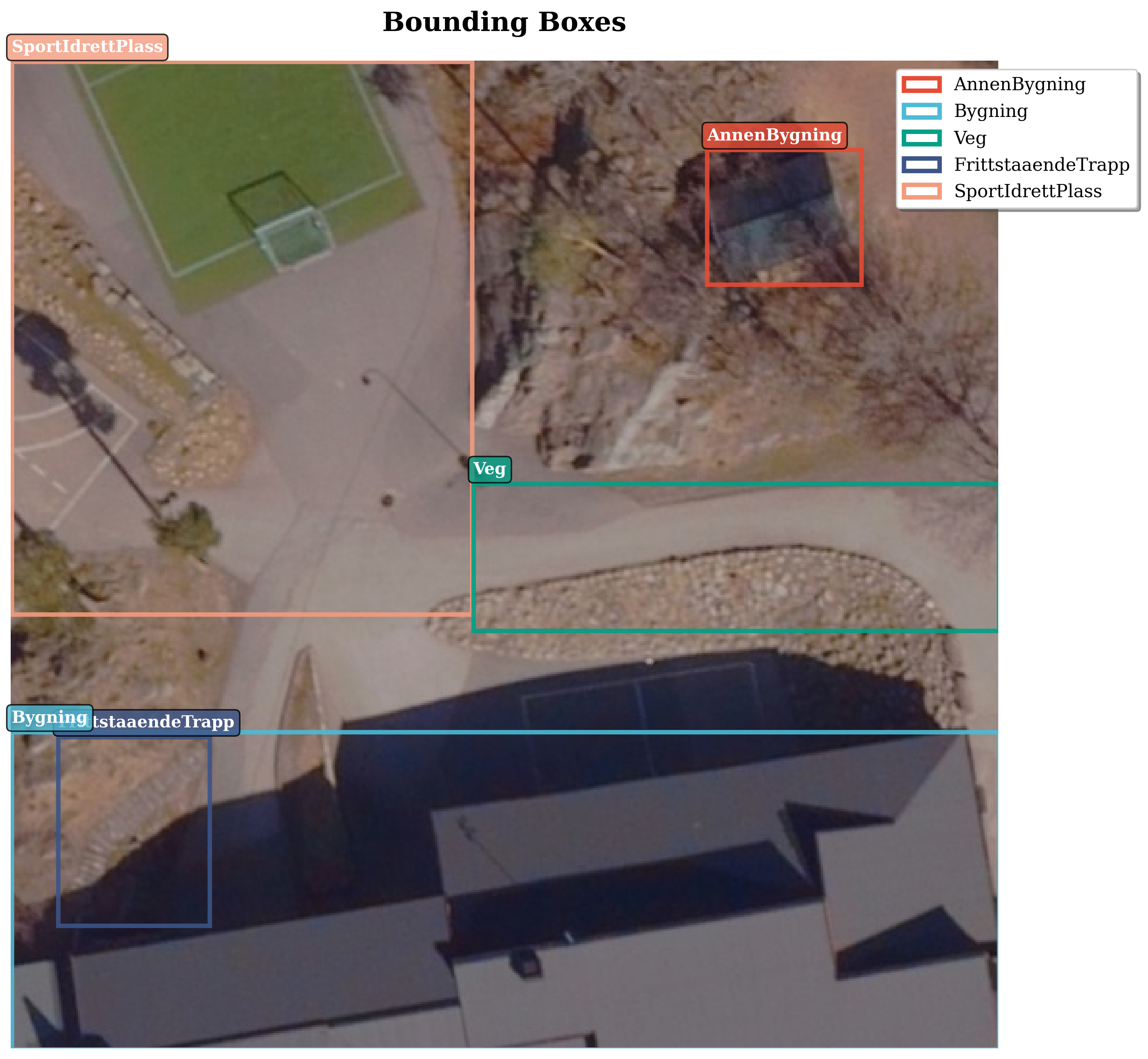} \\
        \includegraphics[width=\textwidth]{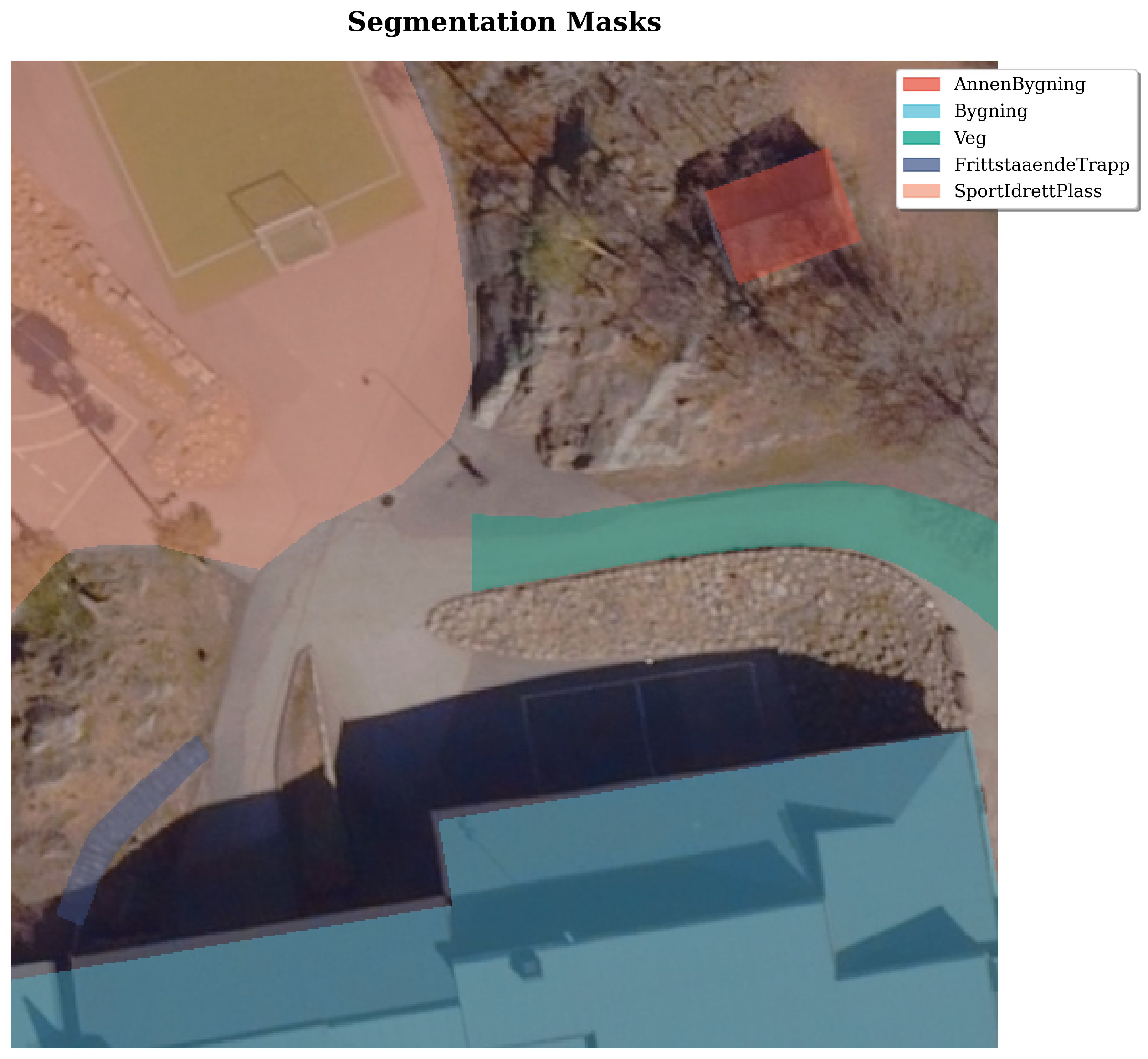}
    \end{minipage}
    \hfill
    \begin{minipage}[t]{0.45\textwidth}
        \centering
        \includegraphics[width=\textwidth]{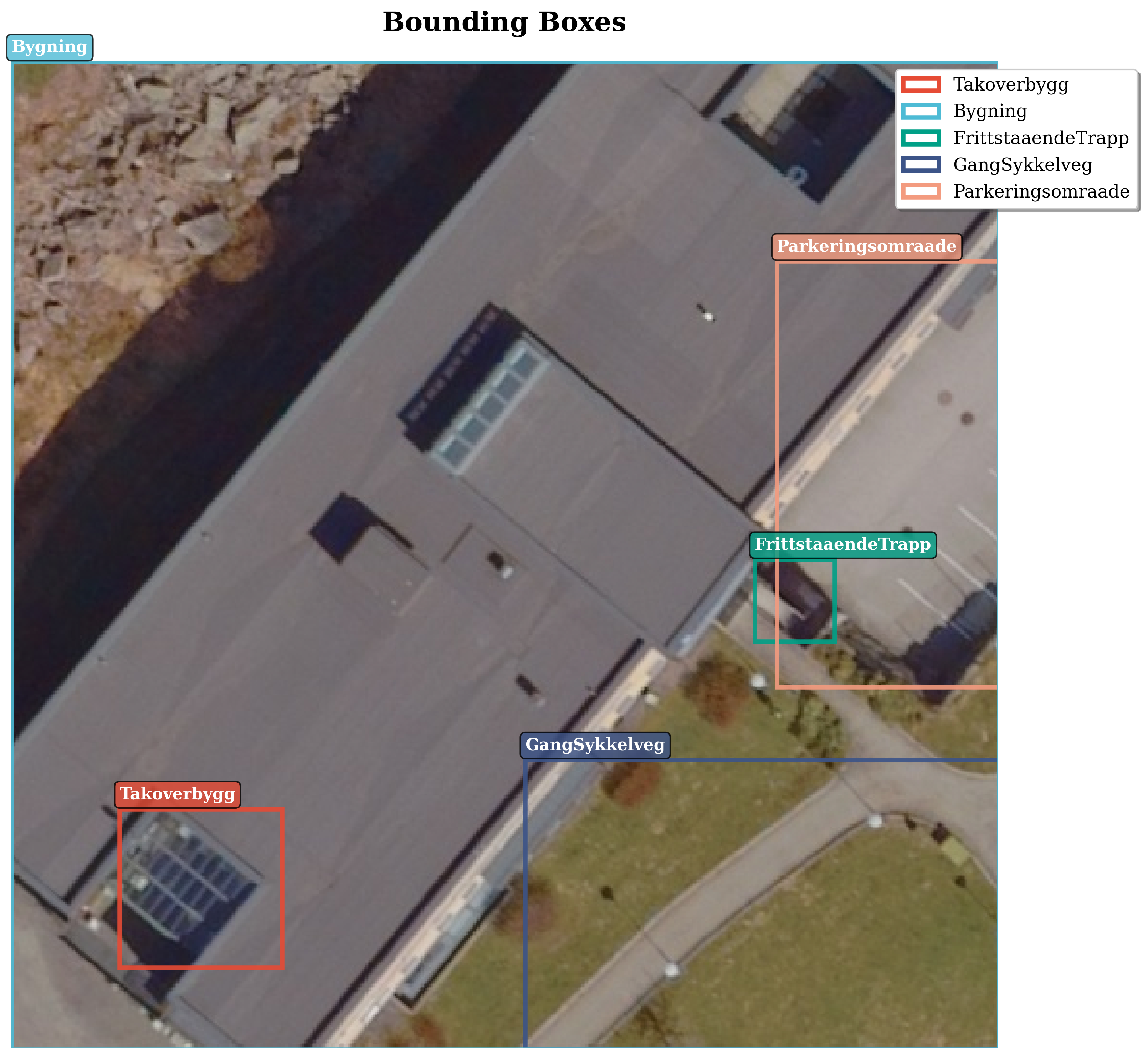} \\
        \includegraphics[width=\textwidth]{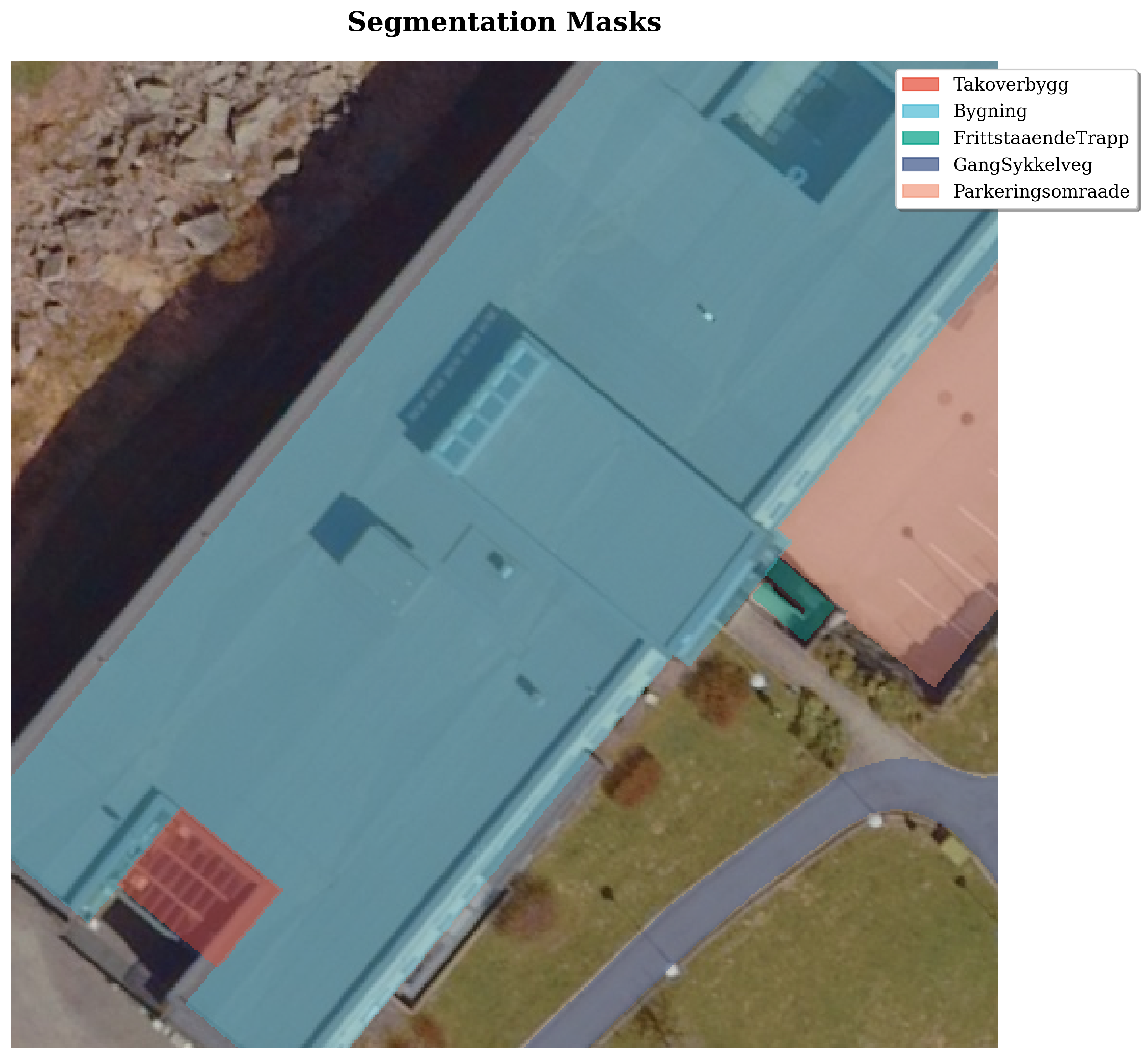}
    \end{minipage}
    \caption{Examples from two locations in the NordFKB dataset. For each location, the top image shows bounding boxes and the bottom image shows the corresponding segmentation mask.}
    \label{fig:dataset_examples}
\end{figure}

\begin{table}[htbp]
\centering
\begin{tabular}{|l|r|r|r|r|r|}
\hline
\textbf{Area} & \textbf{Images} & \textbf{Annotations} & \textbf{Categories} & \textbf{Avg Ann/Img} & \textbf{Avg Classes/Img} \\
\hline
Bergen & 1,379 & 10,027 & 22 & 7.3 & 3.0 \\
Kristiansand & 1,082 & 7,917 & 23 & 7.3 & 3.3 \\
Mo i Rana & 914 & 5,339 & 15 & 5.8 & 2.6 \\
Sandvika & 926 & 3,284 & 25 & 3.5 & 2.3 \\
Stavanger & 4,771 & 12,125 & 24 & 2.5 & 2.0 \\
Tromsø & 1,442 & 8,061 & 26 & 5.6 & 2.8 \\
Verdal & 390 & 838 & 13 & 2.1 & 1.5 \\
\hline
\textbf{Total} & \textbf{10 904} & \textbf{47 591} & \textbf{36} & \textbf{4.9} & \textbf{2.5} \\
\hline
\end{tabular}
\caption{Dataset statistics for the seven areas in the NordFKB dataset.}
\label{tab:area_analysis}
\end{table}

\section{Dataset Structure and Content}

The NordFKB dataset contains 36 distinct semantic classes derived from the official FKB database, representing a broad spectrum of geographic features relevant for mapping, land administration, and planning. Each image tile is provided in \textit{GeoTIFF} format as an orthophoto. The corresponding semantic segmentation annotations are stored as separate binary GeoTIFF mask files, one per class present in the tile. For example, if an orthophoto contains five classes, it is accompanied by five mask files, each containing a binary raster (values 0/1) indicating the pixels belonging to that specific feature type. This per-class design simplifies class-specific loading, processing, and evaluation, and allows the masks to retain full geospatial metadata such as CRS and pixel resolution.

In addition to segmentation masks, the dataset provides object-level bounding box annotations stored in a \textit{JSON} file following the COCO format \cite{coco}, enabling broad compatibility with existing computer vision frameworks and evaluation tools. While both annotation types describe the same geographic features, they serve different purposes and rely on distinct data formats. Bounding boxes are represented as axis-aligned rectangles in COCO JSON and provide efficient object-level supervision for detection models. In contrast, the segmentation masks offer pixel-accurate delineation of feature boundaries and are stored as georeferenced GeoTIFFs aligned with the aerial imagery, supporting fine-grained spatial analysis and direct integration with GIS workflows.

Only image tiles that contain at least one annotated object are included in the dataset, ensuring that all samples are relevant for training and evaluation. The training and validation splits are generated by randomly sampling from all available tiles across the seven geographic areas, rather than splitting by area. This ensures that both splits contain a representative mix of geographic contexts and object types, facilitating fair and robust model comparisons.

Table~\ref{tab:class_distribution_en} shows the distribution of instances for each of the 36 classes \cite{produkt_spesifikasjoner} across the combined dataset, as well as for the training and validation splits. The dataset exhibits a natural class imbalance that reflects real-world geographic distributions, with some classes (e.g., \textit{Bygning} and \textit{VegKjoerende}) occurring frequently, while others (e.g., \textit{Parkdetalj} or \textit{Grustak}) being relatively rare. This diversity allows the benchmark to test both model performance on common features and robustness to infrequent classes.

\begin{table}[htbp]
\centering
\begin{tabular}{|l|l|r|r|r|r|}
\hline
\textbf{Class (NO)} & \textbf{Class (EN)} & \textbf{Train} & \textbf{Val} & \textbf{Train/Val} & \textbf{Combined} \\
\hline
Bygning & Building & 16,195 & 4,170 & 3.88 & 20,365  \\
VegKjørende & Road (Motor traffic) & 6,604 & 1,717 & 3.85 & 8,321 \\
VegGåendeOgSyklende & Road (Pedestrian/Bicycle) & 3,119 & 830 & 3.76 & 3,949 \\
AnnenBygning & Other Building & 1,980 & 508 & 3.90 & 2,488 \\
Takoverbygg & Roofed Structure & 1,877 & 493 & 3.81 & 2,370 \\
Trapp & Stairs & 1,631 & 426 & 3.83 & 2,057 \\
Veg & Road & 1,109 & 243 & 4.56 & 1,352 \\
Parkeringsområde & Parking Area & 768 & 210 & 3.66 & 978 \\
Havflate & Sea Surface & 580 & 131 & 4.43 & 711 \\
Trafikkøy & Traffic Island & 495 & 126 & 3.93 & 621 \\
FrittståendeTrapp & Freestanding Stairs & 440 & 105 & 4.19 & 545 \\
Lekeplass & Playground & 432 & 104 & 4.15 & 536 \\
GangSykkelveg & Pedestrian/Bike Path & 404 & 88 & 4.59 & 492 \\
SportIdrettPlass & Sports Field & 381 & 98 & 3.89 & 479 \\
KaiBrygge & Quay / Pier & 306 & 74 & 4.14 & 380 \\
Anleggsområde & Construction Site & 272 & 70 & 3.89 & 342 \\
Bru & Bridge & 261 & 62 & 4.21 & 323 \\
Elv & River & 217 & 69 & 3.14 & 286 \\
Kanal & Canal & 174 & 29 & 6.00 & 203 \\
Industriområde & Industrial Area & 144 & 36 & 4.00 & 180 \\
Flytebrygge & Floating Dock & 97 & 30 & 3.23 & 127 \\
Innsjø & Lake & 64 & 17 & 3.76 & 81 \\
Tank & Tank & 38 & 16 & 2.38 & 54 \\
SkråForstøtningsmur & Sloped Retaining Wall & 50 & 4 & 12.50 & 54 \\
Park & Park & 36 & 15 & 2.40 & 51 \\
Stein & Rock & 37 & 11 & 3.36 & 48 \\
Svømmebasseng & Swimming Pool & 37 & 11 & 3.36 & 48 \\
Campingplass & Campsite & 26 & 6 & 4.33 & 32 \\
Gravplass & Cemetery & 23 & 5 & 4.60 & 28 \\
Fundament & Foundation & 18 & 7 & 2.57 & 25 \\
Molo & Breakwater / Jetty & 16 & 6 & 2.67 & 22 \\
Tribune & Grandstand & 17 & 3 & 5.67 & 20 \\
ElvBekk & Stream / Creek & 7 & 1 & 7.00 & 8 \\
Pipe & Chimney / Pipe & 5 & 1 & 5.00 & 6 \\
Grustak & Gravel Pit & 4 & 1 & 4.00 & 5 \\
Parkdetalj & Park Detail & 2 & 1 & 2.00 & 3 \\
\hline
\textbf{TOTAL} &  & \textbf{37,866} & \textbf{9,725} & \textbf{3.89} & \textbf{47,591} \\
\hline
\end{tabular}
\caption{Class distribution across combined, training, and validation datasets with English translations.}
\label{tab:class_distribution_en}
\end{table}

The dataset also includes metadata for each tile, including tile identifiers, geographic coordinates, and coordinate reference system (CRS) information. This ensures compatibility with standard GIS software and facilitates integration with other geospatial datasets for downstream analysis.

In addition to the dataset itself, we provide a dedicated benchmarking repository containing data loaders, visualization tools, and evaluation scripts. The repository will be released publicly on GitHub upon acceptance to preserve anonymity during review. The repository includes ready-to-use code for evaluating models on NordFKB, with standardized protocols for both object detection and semantic segmentation. Evaluation metrics include mean Intersection-over-Union (mIoU) for segmentation and mean Average Precision (mAP) for detection. By releasing these tools alongside the dataset, we ensure that results produced by different methods and research groups remain directly comparable and reproducible.

\section{Applications in Mapping and Planning Domains}

The NordFKB dataset enables a wide range of practical and research-oriented applications within mapping, land administration, and spatial planning. Its combination of high-resolution orthophotos and authoritative, fine-grained annotations makes it suitable for training AI models that can automate the detection and mapping of buildings, roads, infrastructure, and natural features. Such models can support the continuous updating of national map databases by identifying new constructions, infrastructure changes, and alterations to the landscape.

In cartographic workflows, the detailed semantic classes allow thematic maps to be generated directly from imagery, reducing the need for manual digitization. Features such as transportation networks, water bodies, vegetation, and recreational areas can be extracted automatically and styled for different map products or scales.

Beyond mapping, the dataset supports advanced geographical analyses, including monitoring urban growth, detecting land-use change, and analysing spatial patterns across diverse Norwegian regions. These capabilities can inform decision-making in urban and regional planning, environmental monitoring, and infrastructure management.

In planning contexts, accurate and up-to-date object detection and segmentation can be used to assess compliance with zoning regulations, identify potential development areas, and monitor building activity over time. Integrating such AI-derived insights into planning processes enables more timely and informed decision-making when responding to changes in the built and natural environment.

Finally, the dataset and accompanying benchmarking tools provide valuable resources for innovation, education, and training. They allow researchers, students, and professionals to work with real-world geospatial data in developing, testing, and evaluating AI methods, fostering broader adoption of geospatial AI in both academic and operational settings.

\section{Discussion}

The introduction of NordFKB addresses a long-standing gap in the availability of authoritative, fine-grained geospatial datasets for Norway. By pairing high-resolution orthophotos with detailed, multi-class annotations derived from the FKB database, it enables the development and benchmarking of AI models aligned with national mapping standards \cite{produkt_spesifikasjoner}. This ensures that methods trained on NordFKB are directly applicable to operational workflows in Norwegian mapping and planning, where precise feature definitions and consistent data quality are essential.

While the dataset offers clear advantages, it also has limitations that should be considered when interpreting results. The coverage is restricted to a sample from seven geographic areas which, although diverse in climate, topography, and urbanization, do not represent the complete range of diverse Norwegian environments. Furthermore, the natural class imbalance with common features such as \textit{Bygning} and \textit{VegKjørende} appearing far more frequently than rare classes like \textit{Grustak} or \textit{Parkdetalj} may affect model performance, particularly for underrepresented categories.

A further consideration is the variable spatial misalignment between orthophotos and segmentation masks for elevated objects, due to radial displacement. This arises from the ortho-rectification process, which references a digital terrain model. As a result, objects above the terrain, such as buildings, can appear displaced in the imagery relative to their accurate ground footprints. These offsets originate from the imagery rather than the annotations and may influence pixel-level segmentation metrics.

From an ethical standpoint, NordFKB is released as publicly available, authoritative mapping data and does not contain personally identifiable information. Nevertheless, care should be taken when combining it with other datasets to avoid unintended privacy implications, particularly in contexts involving residential areas or sensitive infrastructure.

Looking ahead, several opportunities exist for expanding NordFKB. Extending coverage to include a wider range of urban, rural, and natural environments would improve representativeness. Adding other variants of data, such as LiDAR and true-orthophoto, for the same areas would provide more robustness to misalignment. Incorporating temporal data could support change detection and long-term monitoring, while adding complementary data modalities such as LiDAR or hyperspectral imagery could enable multi-sensor research and novel applications. Including raw captured single images and high accurate 3D vector data annotations would enable novel research on 3D reconstruction of objects from airborne sensor data.

Overall, NordFKB provides a strong foundation for advancing geospatial AI in Norway, while also highlighting methodological considerations and avenues for future dataset development.

\section{Conclusion and Future Work}

NordFKB introduces a fine-grained, authoritative benchmark dataset for geospatial AI in Norway, combining high-resolution orthophotos with detailed, multi-class annotations derived from the national FKB database. Covering 36 semantic classes across seven geographically diverse areas, it enables the development and evaluation of models for tasks such as semantic segmentation, instance segmentation, and object detection, all within the framework of nationally defined mapping standards. The dataset’s manual quality control, standardized evaluation protocols, and accompanying benchmarking tools further enhance its value for both research and operational applications.

In future iterations, we plan to expand NordFKB’s geographic coverage to include additional Norwegian environments, increasing its representativeness and robustness. Temporal data will be considered to support change detection and monitoring tasks, while the integration of complementary data sources, such as LiDAR or multispectral imagery, could open new avenues for multi-modal analysis. Additional benchmark tasks may also be introduced to encourage broader research participation and methodological innovation.

By providing a publicly available, high-quality resource rooted in official authoritative mapping data, NordFKB lays the groundwork for advancing AI-based mapping and spatial analysis in Norway. It has the potential to accelerate research in geomatics, cartography, and planning, and to foster the development of methods that can be directly applied in national and local mapping workflows.

\bibliographystyle{IEEEtran}  
\bibliography{references}

\begin{thebibliography}{10}
\providecommand{\url}[1]{#1}
\csname url@samestyle\endcsname
\providecommand{\newblock}{\relax}
\providecommand{\bibinfo}[2]{#2}
\providecommand{\BIBentrySTDinterwordspacing}{\spaceskip=0pt\relax}
\providecommand{\BIBentryALTinterwordstretchfactor}{4}
\providecommand{\BIBentryALTinterwordspacing}{\spaceskip=\fontdimen2\font plus
\BIBentryALTinterwordstretchfactor\fontdimen3\font minus \fontdimen4\font\relax}
\providecommand{\BIBforeignlanguage}[2]{{%
\expandafter\ifx\csname l@#1\endcsname\relax
\typeout{** WARNING: IEEEtran.bst: No hyphenation pattern has been}%
\typeout{** loaded for the language `#1'. Using the pattern for}%
\typeout{** the default language instead.}%
\else
\language=\csname l@#1\endcsname
\fi
#2}}
\providecommand{\BIBdecl}{\relax}
\BIBdecl

\bibitem{geovekst}
``Geovekst,'' \url{https://www.kartverket.no/geodataarbeid/geovekst}, accessed: 12.08.2025.

\bibitem{mapai}
\BIBentryALTinterwordspacing
S.~Jyhne, M.~Goodwin, P.-A. Andersen, I.~Oveland, A.~S. Nossum, M.~Ørstavik, K.~Ormseth, and A.~Flatman, ``Mapai: Precision in building segmentation,'' \emph{Nordic Machine Intelligence}, vol.~2, no.~3, p. 1–3, Sep. 2022. [Online]. Available: \url{http://dx.doi.org/10.5617/nmi.9849}
\BIBentrySTDinterwordspacing

\bibitem{kartai}
``Kartai,'' \url{https://kartai.no/rapporter-og-resultater/}, accessed: 13.08.2025.

\bibitem{ab2_bygg}
KartAi, ``Rapport – kartai, arbeidsgruppe 2: Bygningsidentifikasjon,'' \url{https://kartai.no/wp-content/uploads/2022/02/AP2-Rapport-Byggidentifikasjon-KartAI-17.12.2021-1.pdf}, accessed: 13.08.2025.

\bibitem{optimalisering_ai_analyse}
``Optimalisering avdatasett for bruk i ai-analyse,'' \url{https://kartai.no/wp-content/uploads/2022/02/AP1.2-optimalisering-av-datasett-studentoppgave-1_Gr23-1.pdf}, accessed: 13.08.2025.

\bibitem{pålitelighetsmål}
``Pålitelighetsmål av bygnings-deteksjoner i kartai,'' \url{https://kartai.no/wp-content/uploads/2022/06/AP2.2-Bachelorprosjekt-Var-2022-Palitelighetsmal-av-bygnings-deteksjoner-i-KartAi.pdf}.

\bibitem{kartai_effektivisering}
``Kartai: Effektivisering av byggesaker med kunstig intelligens,'' \url{https://kartai.no/wp-content/uploads/2023/10/nossum-gyland-2023-kartai-effektivisering-av-byggesaker-med-kunstig-intelligens.pdf}, accessed: 13.08.2025.

\bibitem{automatisering}
``Automatisering av dataflyt fra nasjonale kartbaser til ai-algoritmer,'' \url{https://kartai.no/wp-content/uploads/2023/05/AP2.3-Bachelorprosjekt-Var-2023.pdf}, accessed: 13.08.2025.

\bibitem{tilgjengeliggjøring}
``Tilgjengeliggjøring av nasjonale kartdata til bruk i ai-algoritmer,'' \url{https://kartai.no/wp-content/uploads/2024/06/TilgjengeliggjoringavnasjonalekartdatatilbrukiAI-algoritmer-1.pdf}, accessed: 13.08.2025.

\bibitem{kvalitetskontroll}
``Ai-based approaches for quality control of map data,'' \url{https://kartai.no/wp-content/uploads/2025/06/Masteroppgave-Jakob.pdf}, accessed: 13.08.2025.

\bibitem{bygningskontroll}
``Assessing completeness of building objects using yolov8-seg and mask r-cnn,'' \url{https://kartai.no/wp-content/uploads/2025/06/Master-Marianne.pdf}, accessed: 13.08.2025.

\bibitem{spacenet}
\BIBentryALTinterwordspacing
A.~V. Etten, D.~Lindenbaum, and T.~M. Bacastow, ``Spacenet: {A} remote sensing dataset and challenge series,'' \emph{CoRR}, vol. abs/1807.01232, 2018. [Online]. Available: \url{http://arxiv.org/abs/1807.01232}
\BIBentrySTDinterwordspacing

\bibitem{deepglobe}
\BIBentryALTinterwordspacing
I.~Demir, K.~Koperski, D.~Lindenbaum, G.~Pang, J.~Huang, S.~Basu, F.~Hughes, D.~Tuia, and R.~Raskar, ``Deepglobe 2018: {A} challenge to parse the earth through satellite images,'' \emph{CoRR}, vol. abs/1805.06561, 2018. [Online]. Available: \url{http://arxiv.org/abs/1805.06561}
\BIBentrySTDinterwordspacing

\bibitem{inria}
E.~Maggiori, Y.~Tarabalka, G.~Charpiat, and P.~Alliez, ``Can semantic labeling methods generalize to any city? the inria aerial image labeling benchmark,'' in \emph{IEEE International Geoscience and Remote Sensing Symposium (IGARSS)}.\hskip 1em plus 0.5em minus 0.4em\relax IEEE, 2017.

\bibitem{Wang2021_LoveDA}
\BIBentryALTinterwordspacing
J.~Wang, Z.~Zheng, A.~Ma, X.~Lu, and Y.~Zhong, ``Loveda: A remote sensing land-cover dataset for domain adaptive semantic segmentation,'' in \emph{Proceedings of the Neural Information Processing Systems (NeurIPS) Track on Datasets and Benchmarks}, vol.~1.\hskip 1em plus 0.5em minus 0.4em\relax Curran Associates, Inc., 2021. [Online]. Available: \url{https://datasets-benchmarks-proceedings.neurips.cc/paper_files/paper/2021/file/4e732ced3463d06de0ca9a15b6153677-Paper-round2.pdf}
\BIBentrySTDinterwordspacing

\bibitem{Sumbul2019_BigEarthNet}
\BIBentryALTinterwordspacing
G.~Sumbul, M.~Charfuelan, B.~Demir, and V.~Markl, ``Bigearthnet: A large-scale benchmark archive for remote sensing image understanding,'' in \emph{IEEE International Geoscience and Remote Sensing Symposium (IGARSS)}.\hskip 1em plus 0.5em minus 0.4em\relax IEEE, 2019, pp. 5901--5904. [Online]. Available: \url{https://bigearth.net}
\BIBentrySTDinterwordspacing

\bibitem{Yokoya2020_DFC2020}
\BIBentryALTinterwordspacing
N.~Yokoya, P.~Ghamisi, G.-S. Xia, M.~J. Wagner \emph{et~al.}, ``Open data for global multimodal land use classification: Outcome of the 2020 ieee grss data fusion contest,'' in \emph{IEEE International Geoscience and Remote Sensing Symposium (IGARSS)}.\hskip 1em plus 0.5em minus 0.4em\relax IEEE, 2020, pp. 5771--5774. [Online]. Available: \url{https://ieee-dataport.org/open-access/2020-ieee-grss-data-fusion-contest}
\BIBentrySTDinterwordspacing

\bibitem{Xia2023_OpenEarthMap}
\BIBentryALTinterwordspacing
J.~Xia, N.~Yokoya, B.~Adriano, and C.~Broni‐Bediako, ``Openearthmap: A benchmark dataset for global high-resolution land cover mapping,'' in \emph{Proceedings of the IEEE/CVF Winter Conference on Applications of Computer Vision (WACV)}, January 2023, pp. 6254--6264. [Online]. Available: \url{https://openaccess.thecvf.com/content/WACV2023/html/Xia_OpenEarthMap_A_Benchmark_Dataset_for_Global_High-Resolution_Land_Cover_Mapping_WACV_2023_paper.html}
\BIBentrySTDinterwordspacing

\bibitem{kartverket}
``Kartverket,'' \url{https://www.kartverket.no/en/about-kartverket}, accessed: 11.08.2025.

\bibitem{svv}
``Statens vegvesen,'' \url{https://www.vegvesen.no/om-oss/om-organisasjonen/om-statens-vegvesen/}, accessed: 11.08.2025.

\bibitem{nve}
``Noregs vassdrags- og energidirektorat,'' \url{https://www.nve.no/om-nve}, accessed: 11.08.2025.

\bibitem{banenor}
``Bane nor,'' \url{https://www.banenor.no/om-bane-nor/}, accessed: 11.08.2025.

\bibitem{telenor}
``Telenor,'' \url{https://www.telenor.no/om/}, accessed: 11.08.2025.

\bibitem{coco}
\BIBentryALTinterwordspacing
T.~Lin, M.~Maire, S.~J. Belongie, L.~D. Bourdev, R.~B. Girshick, J.~Hays, P.~Perona, D.~Ramanan, P.~Doll{\'{a}}r, and C.~L. Zitnick, ``Microsoft {COCO:} common objects in context,'' \emph{CoRR}, vol. abs/1405.0312, 2014. [Online]. Available: \url{http://arxiv.org/abs/1405.0312}
\BIBentrySTDinterwordspacing

\bibitem{produkt_spesifikasjoner}
``Fkb produkt spesifikasjoner,'' \url{https://www.kartverket.no/geodataarbeid/geovekst/fkb-produktspesifikasjoner}, accessed: 14.08.2025.

\end{thebibliography}

\end{document}